\documentclass[letterpaper, 10 pt, conference,final]{ieeeconf}
\usepackage{graphicx}
\usepackage{amssymb}
\usepackage{amsmath}
\usepackage{bm}
\usepackage{cite}
\usepackage{comment}
\usepackage{color}
\usepackage{url}
\usepackage{alltt}
\usepackage{cleveref}
\usepackage{tikz}
\usepackage{pgfplots}
\usepackage{pgf-umlsd}
\usepackage{cprotect}
\usepackage{multirow}
\usepackage{algorithm}
\usepackage{paralist}
\usepackage{subfigure}
\usepackage[noend]{algpseudocode}
\usepackage{bigfoot}
\usepackage[whole]{bxcjkjatype}
\usepackage{threeparttable}
\usepackage{siunitx}
\usepackage{flushend}

\usetikzlibrary{arrows,shadows,positioning,automata, calc, shapes.misc}
\IEEEoverridecommandlockouts

\def\etal{\textit{et al.}}

\newcommand{\todo}[1]{\textcolor{red}{TODO[#1] }}

\title{\LARGE \bf
Map-based Multi-Policy Reinforcement Learning: \\Enhancing Adaptability of Robots by Deep Reinforcement Learning
}

\author{Ayaka Kume, Eiichi Matsumoto, Kuniyuki Takahashi, Wilson Ko and Jethro Tan
\thanks{All authors are associated with Preferred Networks, Inc., Tokyo, Japan.{\tt\small (e-mail:\{kume, matsumoto, takahashi, wko, jettan\}@preferred.jp)}}%
}

\begin{document}

\maketitle
\thispagestyle{empty}
\pagestyle{empty}

\begin{abstract}
In order for robots to perform mission-critical tasks, it is essential that they are able to quickly adapt to changes in their environment as well as to injuries and or other bodily changes.
Deep reinforcement learning has been shown to be successful in training robot control policies for operation in complex environments. However, existing methods typically employ only a single policy. This can limit the adaptability since a large environmental modification might require a completely different behavior compared to the learning environment.
To solve this problem, we propose Map-based Multi-Policy Reinforcement Learning (MMPRL),
which aims to search and store multiple policies that encode different behavioral features while maximizing the expected reward in advance of the environment change. Thanks to these
policies, which are stored into a multi-dimensional discrete map according to its behavioral feature, adaptation can be performed within reasonable time without retraining the robot. An appropriate pre-trained policy from the map can be recalled using Bayesian optimization. 
Our experiments show that MMPRL enables robots to quickly adapt to large changes without requiring any prior knowledge on the type of injuries that could occur. \par
A highlight of the learned behaviors can be found here: \url{https://youtu.be/QwInbilXNOE}.
\end{abstract}

\section{Introduction}
Humans and animals are well-versed in quickly adapting to changes in not only their surrounding environments, but also to changes to their own body, through previous experiences and information from their senses.
Some example scenarios where such adaptation to environment changes takes place are walking in a highly crowded scene with a lot of other people and objects, walking on uneven terrain, or walking against a strong wind.
On the other hand, examples of bodily changes could be wounds, incapability to use certain body parts due to task constraints, or when lifting or holding something heavy.
In a future where robots are omnipresent and used in mission critical tasks, robots are not only expected to adapt to unfamiliar scenarios and disturbances autonomously, but also to recover from adversaries in order to continue and complete their tasks successfully.
Furthermore, taking a long time to recover or adapt may result in mission failure, while external help might not be available or even desirable, for example in search-and-rescue missions.
Therefore, robots need to be able to adapt to changes in both the environment and their own body state, within a limited amount of time.

Recently, deep reinforcement learning (DRL) has been shown to be successful in complex environments with both high-dimensional action and state spaces~\cite{lillicrap2015continuous, mnih2015human}.
The success of these studies relies on a large number of samples in the orders of millions, so re-training the policy after the environment change is unrealistic.
Some methods avoid re-training by increasing the robustness of an acquired policy and thus increasing adaptability.
In robust adversarial RL, for example, an agent is trained to operate in the presence of a destabilizing adversary that applies disturbance forces to the system~\cite{pinto2017robust}.
However, using only a single policy limits the adaptability of the robot to large modifications which requires completely different behaviors compared to its learning environment.
\begin{figure}
\centering
	\subfigure{\includegraphics[width=0.97\columnwidth]{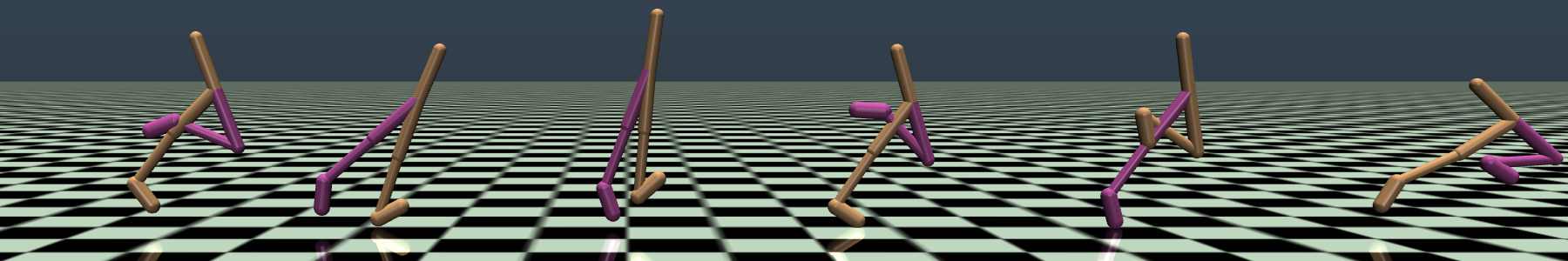}}
    \subfigure{\includegraphics[width=0.97\columnwidth]{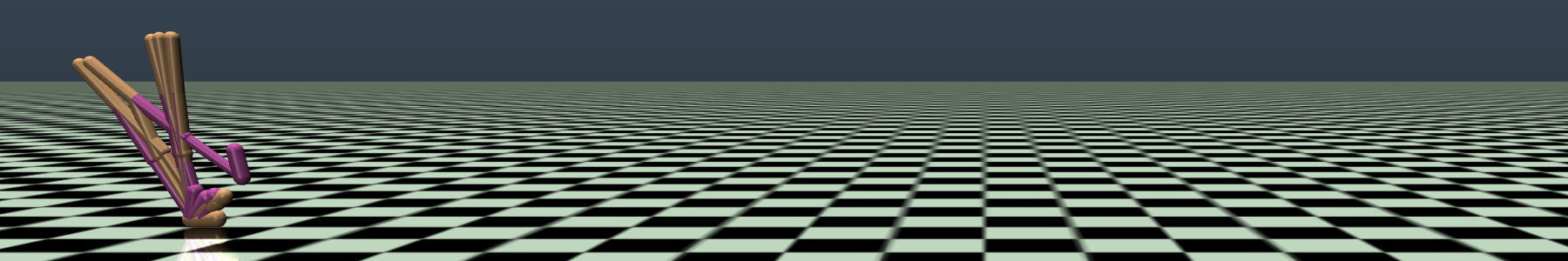}}
	\subfigure{\includegraphics[width=0.97\columnwidth]{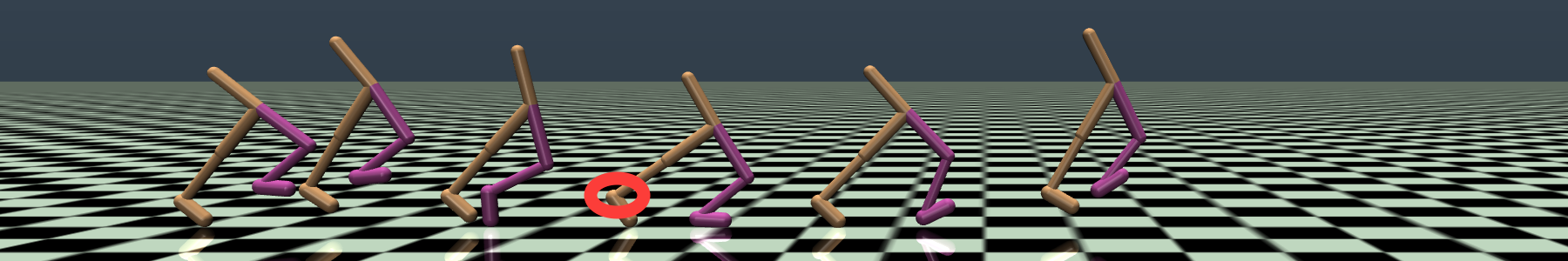}}
	\caption{Time lapse of the OpenAI Walker2D model walking for 360 time steps using a policy and succeeding while intact (top),
	failing due to a joint being limited (middle),
	and succeeding again post-adaptation despite the limited joint marked in red by selecting an appropriate policy using our proposed method (bottom).}
	\label{fig:overview}
	\vspace{-5mm}
\end{figure}

We propose Map-based Multi-Policy Reinforcement Learning (MMPRL), which trains many different policies by combining DRL and the idea of using a behavior-performance map~\cite{cully2014robots}.
MMPRL aims to search and store multiple possible policies which have different behavioral features while maximizing the expected reward in advance in order to adapt to the unknown environment change.
For example, there are various ways for multi-legged robots to move forward: walking, jumping, running, side-walking, etc.
In this example, only the fastest policy would survive when using ordinary RL, whereas MMPRL saves all of them as long as they have different behavioral features.
These policies are stored into a multi-dimensional discrete map according to its behavioral feature.
As a result, adaptation can be done within reasonable time without re-training the robot, but just by searching an appropriate pre-trained policy from the map using an efficient method like Bayesian optimization, see~\Cref{fig:overview}.
We show that, using MMPRL, robots are able to quickly adapt to large changes with little knowledge about what kind of accidents will happen.

The rest of this paper is organized as follows.
Related work is described in~\Cref{sec:relatedwork}, while~\Cref{sec:method} explains our proposed method.
\Cref{sec:experimentsetup} outlines our experiment setup and evaluation setting, with results presented in~\Cref{sec:results}.
\Cref{sec:conclusion} describes our future work and conclusion.

\section{Related Work}
\label{sec:relatedwork}
\subsection{Self-recovering Robots}
A variety of methods have been employed in existing work on self-recovering robots.
Most of this work has focused on minimizing the time needed to adapt.
An intelligent trial-and-error method (IT\&E) which quickly adapts to a large change in the robot state such as an unexpected leg injury has been proposed in~\cite{cully2014robots} and~\cite{chatzilygeroudisreset}.
%
%
A stochastic derivative-free method is used to populate a map of behaviors, from which a policy is selected at the time of damage.
However since this method uses predetermined periodic patterns, it cannot change its gait according to exteroceptive sensor readings of the environment.
Bongard \etal~\cite{bongard2006resilient} dealt with damage recovery by continuously comparing an updated simulation of the robot with a predetermined undamaged self-model in order to track changes and damages to the body.
Following this work, Koos \etal~\cite{koos2013fast} also used a self-model.
The main difference is that this work uses the undamaged self-model of the robot to discover new behaviors, instead of continuously updating it.
This reduces the complexity of the problem since it is not attempted to diagnose damages, making it faster to adapt.
However these works also do not take considerable environmental changes into account.

Yet another approach was proposed in~\cite{ren2015multiple}, which focuses on damage recovery for legged robots implemented with multiple central pattern generator (CPG) controllers.
The CPG controllers are synchronized by default, until a leg is damaged.
A learning algorithm is then employed to learn the remaining legs' oscillation frequencies to resynchronize them, enabling the robot to continue to walk.
However, this work does not seem to focus on adapting quickly.

\subsection{Improving Robustness in Deep Reinforcement Learning}
Deep reinforcement learning (DRL) has recently gained much attention and its range of applications has extended to the fields of game playing \cite{mnih2015human,mnih2016asynchronous}, robot control \cite{lillicrap2015continuous,schulman2015trust,levine2016end}, natural language \cite{li2016nlp}, neural architecture search \cite{zoph2017} etc.
For more details, we kindly refer the reader to the survey \cite{li2017deep}.
One of the important problem settings in deep reinforcement learning is domain adaptation: training an agent to adapt to a new environment different from its training environment within a short amount of time (e.g., knowledge transfer from simulation to a real setting, compensating for physical parameter changes, grabbing unknown objects, etc.).

Various approaches exist to tackle this problem, such as estimating system parameters in the new environment online~\cite{yu2017preparing, Heess2015},
training a prediction model with a few examples and using it to change the policy~\cite{higuera2017adapting, williams2017information},
source domain randomization during training to improve robustness~\cite{tobin2017domain, rajeswaran2016epopt},
adversarial training~\cite{pinto2017robust, pinto2017supervision}, and introducing modularity or hierarchy~\cite{pmlr-v70-vezhnevets17a} to encourage the reuse of submodules corresponding to unchanged parts in the environment.

\subsection{Neuroevolution}
\label{ss:neuro_evolution}
Another area relevant to our research is neuroevolution, which is a technique to optimize neural networks with genetic methods, and has been studied extensively \cite{yao1993review, angeline1994evolutionary, yao1999evolving}.

Increasing the size of the networks in these methods has been challenging, because the search process relies on random perturbations on these parameters, making it vulnerable to the curse of dimensionality.
Recently, researchers have applied neuroevolution to deep neural networks consisting of many parameters.
For instance, Miikkulainen \etal\cite{miikkulainen2017evolving} used genetic methods to search layer connections in combination with supervised learning to weight optimization, while Alvernaz \etal\cite{alvernaz2017autoencoder} have reduced the number of dimensions of a visual input by using an auto-encoder and applied genetic methods on compressed hidden states.
Another remarkable work has been done by Salimans \etal in~\cite{salimans2017evolution}, in which they propose an evolutionary strategy while getting results comparable to state-of-the-art deep reinforcement learning methods with the same network architecture, though requiring several times the number of samples for complex environments.


\section{Proposed method}
\label{sec:method}
%
%
We propose MMPRL, a method which combines intelligent trial-and-error (IT\&E)~\cite{cully2014robots} with deep reinforcement learning (DRL) to make a robust system that deals with bodily and environmental changes within a short amount of time. 
Moreover, we improve the sample efficiency in the IT\&E algorithm by changing its search method during the map creation phase.
Specifically, we replace the random mutation step in IT\&E with an update by DRL.
This raises the likelihood to mutate in the correct direction, which in consequence allows us to learn larger networks.
The result is a DRL method that is capable of adapting to both environmental and bodily changes.

We have chosen Deep Deterministic Policy Gradient (DDPG)~\cite{lillicrap2015continuous} as the DRL method, but other methods can also be used.
MMPRL trains thousands of different policies and searches for the appropriate policy among these to adapt to new situations within only a few iterations.
Unlike previous methods, which aim to train a single strong policy, each policy can be trained with a simple network architecture and learning rule, instead of more sophisticated methods such as RNN or hierarchical RL.
The following subsections will briefly discuss the methods our work is based on.

\subsection{Deep Deterministic Policy Gradient (DDPG)}
\label{ssec:ddpg}
Let $(S, A, P, r, \gamma, s_0)$ be a Markov decision process with state $s\in S$, (continuous) action $a\in A$, transition probability $P(s'\vert s, a)$, reward function $r(s, a)\in \mathbb{R}$, discount factor $\gamma \in [0, 1]$, and initial state distribution $s_0$.
RL typically learns a policy $\pi: S\rightarrow A$ to maximize the expected discounted accumulated reward $J = \mathbb{E}[R_1]$ from the initial state, where $R_t = \sum_{i=t}^{T}\gamma^{(i-t)}r(s_i,a_i)$.
DDPG~\cite{lillicrap2015continuous} is a DRL method to deal with multi-dimensional continuous action spaces.
This algorithm learns not only a policy (actor), but also learns the state-action value function $Q(s, a\vert \theta^Q)$ (critic).
The critic is learned as in Q-learning, by minimizing the temporal difference error derived from the Bellman equation~\cite{watkins1992q}.
But calculating the maximum of the value function $\mathrm{max}_a Q(s, a)$ is computationally expensive, so it is replaced by the policy function output $\mu(s)$:

\begin{equation}
\label{eq:critic}
\begin{split}
L(\theta^Q) = \mathbb{E}[(Q(s,a\vert \theta^Q) - y)^2] \\
y = r\ +\ \gamma Q(s', \mu(s'\vert \theta^\mu)\vert \theta^Q)
\end{split}
\end{equation}
where $(s, a, s', r)$ is the one-step transition (state, action, next state, reward) taken from the agent's experience.
The actor is updated as shown in ~\Cref{eq:grad}, which steers the output in the direction of the gradient to maximize the output of the critic.

\begin{equation}
\label{eq:grad}
\nabla_{\theta^\mu} J \sim \mathbb{E}[\nabla_{a} Q(s, a\vert \theta^Q) \nabla_{\theta^\mu} \mu(s|\theta^\mu)]
\end{equation}

Like in Deep Q-learning (DQN)~\cite{mnih2013playing}, a replay buffer is used to sample a minibatch to minimize correlations between samples.
The minibatch is then used to update both the actor and critic.

\subsection{Intelligent Trial and Error (IT\&E)}
IT\&E is an algorithm proposed by Cully ~\etal~\cite{cully2014robots}, which is based on evolutionary strategies, to quickly adapt to abnormalities.
This method consists of two phases: the map creation phase (MAP-Elites algorithm) and the adaptation phase (M-BOA algorithm).
In the map creation phase, many policies are stored in a high dimensional map which represents features of a specific task. 
When an unexpected situation occurs, the default policy might not work any longer.
In that case, Bayesian optimization~\cite{mockus2012bayesian} is used to search the map for a policy more suited to the new situation.

\subsubsection{Map creation phase}
The map used in this method is multidimensional, and at most one policy and its performance is stored at each coordinate.
Each coordinate can be accessed through a behavioral descriptor, which describes the general characteristics of various possible policies and is designed by the user.
This means that policies with coordinates close to each other, also have a similar behavior.
In the initialization step policies are generated randomly, for which the performance and behavioral descriptor is evaluated.
The behavioral descriptor is then used to store the generated policy in the map, or to replace another less performing one if the coordinate is already occupied.

Once the initialization step is finished, the algorithm will update the policies similar to population based methods such as evolutionary strategies.
This way, a policy is able to mutate such that its behavioral descriptor transforms, which can then again be stored in a new location in the map.
We kindly refer the reader to~\cite{cully2014robots} for more details.

\subsubsection{Adaptation phase}
In the adaptation phase, the created behavior-performance map is explored to find a policy that suits the current situation using Bayesian optimization.
During this phase, performance of the robot is repeatedly observed and used to re-evaluate the performance of the stored policies.
After this update, the algorithm will continue to search for a new policy until one is found that works in the new situation.

Gaussian Processes are used to re-evaluate the performance of the policies stored in the map $\mathcal{P}(\bm{\mathrm{x}})$, after a change has occurred.
A Mat\'{e}rn kernel was used as the kernel function $k(\bm{\mathrm{x}}, \bm{\mathrm{y}})$.
Given observations $\{(\chi_i, P_i)\}, i=1\dots t$, where $\chi_i$ is the selected coordinate of the map at the iteration $i$ and $P_i$ is an observed performance with policy which is stored at $\chi_i$, the estimation of performance at $\bm{\mathrm{x}}$ is a Gaussian distribution $\mathcal{N}(\mu_t(\bm{\mathrm{x}}), \sigma_t^2(\bm{\mathrm{x}}))$, where
\begin{equation}
\begin{split}
&\mu_t(\bm{\mathrm{x}}) = \mathcal{P}(\bm{\mathrm{x}}) + \bm{\mathrm{k}}^T\bm{\mathrm{K}}^{-1}(\bm{\mathrm{P}}_{1:t} - \mathcal{P}(\chi_{1:t})) \\
&\sigma_t^2(\bm{\mathrm{x}}) = k(\bm{\mathrm{x}}, \bm{\mathrm{x}}) - \bm{\mathrm{k}}^T\bm{\mathrm{K}}^{-1}\bm{\mathrm{k}}\\
&\bm{\mathrm{K}} = \begin{bmatrix}
k(\chi_1,\chi_1) &  \cdots & k(\chi_1,\chi_t) \\
\vdots &  \ddots & \vdots \\
k(\chi_t,\chi_1) &  \cdots & k(\chi_t,\chi_t)
\end{bmatrix}
+ \sigma_{noise}^2 I\\
&\bm{\mathrm{k}} = [k(\bm{\mathrm{x}}, \chi_1), k(\bm{\mathrm{x}}, \chi_2), \cdots k(\bm{\mathrm{x}}, \chi_t)]
\end{split}
\end{equation}
$\sigma_{noise}^2$ is an user defined parameter that model the observation noise.

The next observation is selected by taking the maximum of the upper confidence bound:
\begin{equation}
\chi_{t+1} \leftarrow \mathrm{arg max}_{\bm{\mathrm{x}}}
(\mu_t(\bm{\mathrm{x}})
+\kappa\sigma_t(\bm{\mathrm{x}}))
\end{equation}

Once again, we refer the reader to \cite{cully2014robots} for more detailed information on the entire adaptation phase.

\subsection{Map-based Multi-Policy Reinforcement Learning}
\label{ssec:mmprl}
It is well known that in DRL algorithms, including DDPG, performance usually depends on randomness in the environment, the exploration and furthermore the initialization of the weights, as shown in \cite{islam2017reproducibility}.
This causes the results to vary, which is usually not desirable.
However, this also means that whenever DDPG is unstable, a new policy is born with a different behavior.
In our case, this can be utilized to generate policies with very different behaviors, which can then be stored in a map as done in IT\&E.

Like IT\&E, our method has a map creation phase and an adaptation phase.
Time is abundant before deployment of the robot, so we choose to perform the map creation phase during this time to store as many policies as possible, while the adaptation phase will start when needed during task execution.
Note that our work does not actually detect when the robot should transit to the adaptation phase from its normal phase, meaning that we do not detect the change itself to which the robot needs to adapt.
We believe that this can be easily done using various existing methods, but we consider this to be out of our scope.

For our adaptation phase we use M-BOA, which is the same as in the original IT\&E algorithm.
However our map creation phase is a modified version of the original MAP-Elites algorithm.
In the map creation phase of IT\&E, new behaviors are generated by random perturbation of existing behaviors, so this method is only effective in low-dimensional parameter spaces~\cite{cully2014robots}.
In our map creation phase, we replace the random perturbation part by an update of DDPG.
By making use of DDPG and its instability, we can generate new policies despite of using high-dimensional parameter spaces.
We use multiple DDPG agents configured with different initial values, to stimulate the generation of policies with varying behaviors and to parallelize the system.
Furthermore, each agent uses its own replay buffer at all times.
The result is increased stability of the total system, despite the high variation in the learning result of each agent.

\Cref{alg:map_elite} shows the outline of the algorithm for one agent in the map creation phase.
Each agent updates the actor and critic with DDPG until its iteration reaches $I_{\mathrm{init}}$ according to~\Cref{eq:critic} and~\Cref{eq:grad} as described in~\Cref{ssec:ddpg}.
For each iteration, we measure the performance of the policies and update the behavior-performance map similar to IT\&E.
Both the actor and critic used in the iteration are stored in the map.
After $I_{\mathrm{init}}$ iterations of the algorithm, we randomly pick a stored pair of policy and critic every once in \textsc{freq} iterations from the map.

\algdef{SE}[DOWHILE]{Do}{doWhile}{\algorithmicdo}[1]{\algorithmicwhile\ #1}%
\begin{algorithm}
\caption{MMPRL (single agent)}
\label{alg:map_elite}
\begin{algorithmic}
\State $(\mathcal{P} \gets \emptyset, \mathcal{C} \gets \emptyset)$ \Comment{Initialize a behavior-performance map.}
\State Initialize policy $\mu^{\prime}$, critic $Q^{\prime}$, replay buffer $\mathcal{R}$.
\For{$\rm{\textsc{iter}}=1$ to $I$}
\If{$\rm{\textsc{iter}} < I_{init} \rm{\ or\ } (\textsc{iter}\mod\textsc{freq}) \ne 0$}
\State $\rm{\mu, Q} \gets \rm{\mu^{\prime}, Q^{\prime}}$
\Else
\State $\rm{\mu, Q} \gets \rm{random\_selection\_from\_map}(\mathcal{C})$ 
\EndIf
\State $\rm{\mu^{\prime}}, Q^{\prime}, \mathcal{R} \gets \rm{DDPG}(\mu, Q, \mathcal{R})$ 

\State $\rm{x^{\prime}} \gets \rm{behavioral\_descriptor(simu(\mu^{\prime}))}$
\State $\rm{p^{\prime}} \gets \rm{performance(simu(\mu^{\prime}))}$
\If{$\mathcal{P}\rm{(x^{\prime})}=\emptyset \rm{\ or\ } \mathcal{P}\rm{(x^{\prime})}<p^{\prime}$}
\State $\mathcal{P}\rm{(x^{\prime})} \gets p^{\prime}$
\State $\mathcal{C}\rm{(x^{\prime})} \gets \rm{\mu^{\prime}, Q^{\prime}}$
\EndIf
\EndFor
\end{algorithmic}
\end{algorithm}

\section{Experiment Setup}
\label{sec:experimentsetup}

To evaluate MMPRL, we conduct experiments using a simulated environment in OpenAI Gym~\cite{1606.01540} in combination with the MuJoCo 1.50~\cite{todorov2012mujoco} physics engine.
As deep learning framework we use Chainer~\cite{chainer_learningsys2015}.
All our experiments were run on a machine equipped with 128\,GB RAM, an Intel Xeon E5-2523v3 CPU, and a Maxwell GeForce GTX Titan X.

\subsection{Robot Models}
The models we use in simulations are the hexapod robot acquired via~\cite{hexapod_model}, and the Walker2D from OpenAI Gym, see~\Cref{fig:models}.

\subsubsection{Hexapod}
%
Each leg of the hexapod has three actuators. The height of the robot is \SI{0.2}{\metre} in the neutral position.
Its goal is to move forward in the world, which in our case is along the x-axis.
Furthermore, we use target joint angle positions of all \num{18} joints as action, each of which ranges from \numrange{-0.785}{0.785} \si{\radian}.

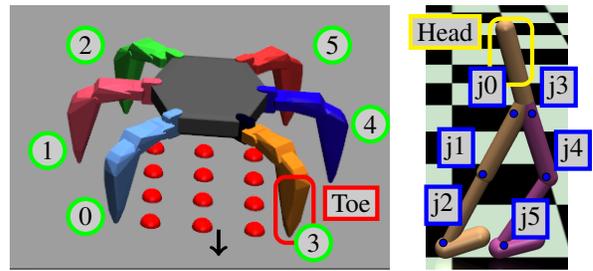
\begin{figure}
\centering
	\vspace{2mm}
	\subfigure{
    \begin{tikzpicture}[every node/.style={scale=0.5, font=\fontsize{20}{22.4}\selectfont}]
 \tikzstyle{htext}=[rectangle,draw=red,ultra thick,inner sep=5pt,fill=black!20]
 \tikzstyle{leg}=[circle,draw=green,ultra thick,fill=black!20]
 \node[anchor=south west,inner sep=0] (image) at (0,0) {\includegraphics[height=7cm]{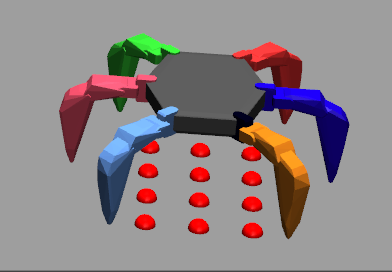}};
 \begin{scope}[x={(image.south east)},y={(image.north west)}]
  \draw[red,ultra thick,rounded corners] (0.70,0.35) rectangle (0.80,0.10);
  \node (toe) at (0.80,0.25) {}; \node[right of=toe,htext] {Toe};
  \node[leg] (l0) at (0.20, 0.20) {0};
  \node[leg] (l1) at (0.10, 0.45) {1};
  \node[leg] (l2) at (0.20, 0.85) {2};
  \node[leg] (l3) at (0.80, 0.10) {3};
  \node[leg] (l4) at (0.95, 0.55) {4};
  \node[leg] (l5) at (0.85, 0.85) {5};
  \draw[arrows=->, line width=0.5mm, draw=black] (0.55,0.15) -- (0.55,0.05);
 \end{scope}

\end{tikzpicture}}
    \subfigure{
 \begin{tikzpicture}[dot/.style={draw,circle,minimum size=2mm,inner sep=0pt,outer sep=0pt,fill=blue}, every node/.style={scale=0.5, font=\fontsize{20}{22.4}\selectfont}]
 \tikzstyle{jtext}=[rectangle,draw=blue,ultra thick,inner sep=5pt,fill=black!20]
 \tikzstyle{htext}=[rectangle,draw=red,ultra thick,inner sep=5pt,fill=black!20]
 \tikzstyle{headtext}=[rectangle,draw=yellow,ultra thick,inner sep=5pt,fill=black!20]
 \node[anchor=south west,inner sep=0] (image) at (0,0) {\includegraphics[height=7cm]{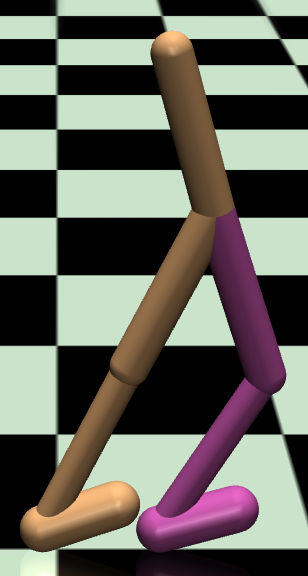}};
 \begin{scope}[x={(image.south east)},y={(image.north west)}]
  \draw[yellow,ultra thick,rounded corners] (0.45,0.95) rectangle (0.75,0.7);
  \node (head) at (0.40,0.90) {}; \node[left of=head,headtext] {Head};
  \node[dot] (j0) at (0.62, 0.59) {}; \node[above left of=j0,jtext] {j0};
  \node[dot] (j1) at (0.40, 0.36) {}; \node[above left of=j1,jtext] {j1};
  \node[dot] (j2) at (0.12, 0.10) {}; \node[above of=j2,jtext] {j2};
  \node[dot] (j3) at (0.75, 0.59) {}; \node[above right of=j3,jtext] {j3};
  \node[dot] (j4) at (0.85, 0.35) {}; \node[above right of=j4,jtext] {j4};
  \node[dot] (j5) at (0.55, 0.09) {}; \node[above right of=j5,jtext] {j5};
 \end{scope}
 \end{tikzpicture}
    }
	\caption{Models used in our experiments: hexapod robot (left), where the numbers represent the leg index and the arrow depicts its forward direction.
	In some experiments we refer to the toe of the robot, which is marked in red here for leg 3.
	The red dots underneath the hexapod represent the grid from which relative height is measured for observation;
	Walker2D (right) with its head marked in yellow (also used in some of our experiments) and the joints are marked by the blue dots.}
\label{fig:models}
\vspace{-5mm}
\end{figure}

For the agent observations, we use a \num{41}-dimensional vector which includes the current height of the robot (1), orientation of the robot in quaternions (4), joint angle positions of all joints in each leg (18), booleans for each leg whether it is making contact with the ground (6), and the relative height of 12 points in space, see~\Cref{fig:models}, which form a grid below the robot (12).

The time step of the hexapod simulation is \SI{0.01}{\second} and an action is selected every \num{5} time steps.
We limit an episode to be at most \num{1000} steps and the episode will end whenever the robot falls down on its legs or jumps above a certain height.

The reward function $R$ is defined as follows.
\begin{equation*}
R_t = \Delta x_t + s_t + w_0 C_t - w_1 \vert \vert \tau_t \vert \vert_2^2 - w_2 \vert \vert \phi_t \vert \vert_2^2
\end{equation*}
where $\Delta x_t$ is the covered distance of the robot in the current time step since the previous time step,
$s_t$ is the survival reward, which is \num{0.1} on survival and \num{0} if the episode is terminated by the aforementioned conditions,
$C_t$ is the number of legs making contact with the ground,
$\tau_t \in \mathbb{R}^{18}$ is the vector of squared sum of external forces and torques on each joint,
$\phi_t \in \mathbb{R}^{18}$ are the target joint angles (the actions),
and $w_n$ is the weight of each component with $w_0 = 0.03$, $w_1 = 0.0005$, and $w_2 = 0.05$.
The values of these weights remain the same throughout all of our experiments.

We use a 6-dimensional map to store the policies, of which the performance is evaluated by the total forward distance in \num{1000} time steps.
Furthermore, we define the behavioral descriptor corresponding to this map as
$[F_{0},F_{1},F_{2},F_{3},F_{4},F_{5}]$,
where $F_{n} \in \{0, 0.25, 0.5, 0.75, 1\}$ represents the fraction of the stance time of leg $n$ during one episode.

\subsubsection{Walker2D}
The Walker2D model has two legs and each leg also contains three joints, see~\Cref{fig:models}.
Again, we use a 6-dimensional map to store policies and performance is evaluated in the same fashion using the total forward distance in \num{1000} time steps.
However, we define a different behavioral descriptor for the Walker2D:
\begin{equation*}[\delta(0, 3), \delta(1, 4), \delta(2, 5), \sigma(0, 3),\sigma(1, 4), \sigma(2, 5)]\end{equation*}

Where
$
\delta(i, j) = \sum_t{\lvert \phi^{j}_{t} - \phi^{i}_{t}\rvert}~\mathrm{,}\enskip~\sigma(i, j) = \sum_t{\lvert \phi^{j}_{t} + \phi^{i}_{t}\rvert}
$.
$\phi^{i}_{t}$ and $\phi^{j}_{t}$ are the angles of joint $i$ and $j$ respectively in time step $t$.
Each entry in the descriptor is quantized uniformly to \num{5} values between the min and max value of the entry.

For more detailed information on the action, observation, and reward function, we refer the reader to OpenAI Gym.

\subsection{Evaluation Settings of Environment and Robots}
\label{ssec:environments}
Both agents are trained with constant settings, while the agents are tested in various environments and conditions in the adaptation phase.
We remind the reader that we do not actually detect when a change has occured, and thus when to initiate the adaptation phase.
To still test the adaptation performance we start the simulation with the changes already applied, to initiate the adaptation phase immediately.

\subsubsection{Hexapod}
\begin{figure}
	\centering
	\vspace{2.2mm}
	\subfigure[]{\includegraphics[height=20mm]{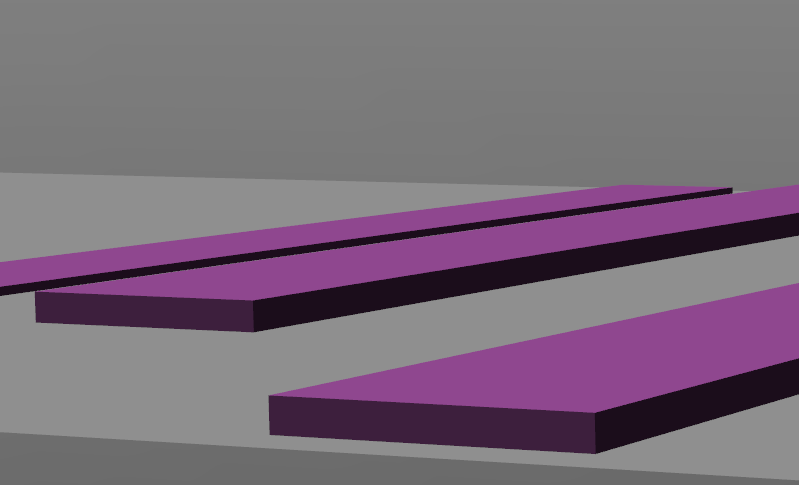}}
    \subfigure[]{\includegraphics[height=20mm]{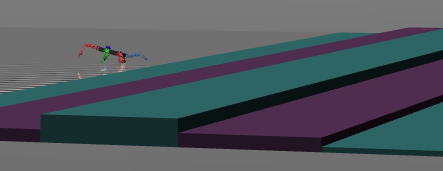}}
	\caption{Obstacles used during the (a) training phase of our hexapod experiments, and (b) adaptation phase in the climbing stairs experiment.}
	\label{fig:plane}
	\vspace{-5mm}
\end{figure}

For the training phase, we trained the hexapod undamaged on the random stairs as shown in~\Cref{fig:plane}.
For the first meter, no obstacles are included in the environment.
After the first meter, some steps are placed as obstacles with random height $\mathcal{H}\sim\mathcal{U}(\num{0.02}, \num{0.1})$\,m, width (y direction) $\mathcal{W}\sim\mathcal{U}(\num{4}, \num{12})$\,m, depth (x direction) $\mathcal{D}\sim\mathcal{U}(\num{0.1}, \num{1.0})$\,m  and location.
Note that the max possible height is \SI{50}{\percent} of the robot height in its neutral position.

We tested the adaptation performance by changing the situation, where we either
\begin{inparaenum}[(i)]
	\item let the robot walk on a flat plane after we replace either one or two toes with links of \SI{1}{\milli\metre},
	\item keep the robot intact and the plane flat, but introduce an observation delay on purpose, or
	\item make the robot climb stairs with step heights different from the training phase.
\end{inparaenum}

In situation (ii) we delay the observation by $N$ time steps such that the observation at time step $t$ becomes the one at $t-N$.
For situation (iii) the robot is undamaged and each step size of the stairs is chosen as $\mathcal{H}\sim\mathcal{U}(-a, a)~\mathrm{where}~a\in\mathbb{R}$.
An example of such a situation is shown in~\Cref{fig:plane} where we sampled the step height with $a = 0.1$.

\subsubsection{Walker2D}
We trained the intact Walker2D on the same environment as provided by OpenAI.
For the adaptation experiment, we tested the agent's performance under the following conditions, where we
\begin{inparaenum}[(i)]\item  limit the control range of one of the joints from $[-1,1]$ to $[-0.01, 0.01]$,
\item change the head radius to $\frac{l}{10}\cdot0.05$ where $l \in \mathbb{R}$, where \SI{0.05}{\metre} is the original head radius,
and \item simulate walking against a slope of $S$ degrees by changing the magnitude and direction of the simulated gravity\end{inparaenum}.

\subsection{Hyperparameters}
\label{ssec:hyperparameters}
We note that the values of the following hyperparameters are found by trial and error during our experiments.
For the hexapod, we use a multilayer perceptron (MLP) with three hidden layers (\num{400}, \num{400}, \num{200}) for the actor network.
We map the action and observation to a linear layer with 100 units (action) and 300 units (observation) for the critic network.
These layers in the critic network are concatenated and connected afterwards to an MLP with two hidden layers (\num{400}, \num{200}).

For the Walker2D we use an MLP with three hidden layers with (\num{40}, \num{40}, \num{20}) units for the actor, and for the critic a linear layer with \num{16} units for the action and \num{48} units for the observation.
The layers in the critic network are concatenated and connected to an MLP, but now with (\num{64}, \num{32}) units.

For all networks in both agents, ReLU is used as activation function in the hidden layers, while a hyperbolic tangent is used for the final layer.
As optimizer, we use Adam~\cite{kingma2014adam} with $\alpha=\num{0.0001}$.
Moreover, we use a discount factor of $\gamma = \num{0.99}$, a soft target update rate of $\tau = \num{0.01}$, and a replay buffer with a size of \num{2000000} for DDPG.
During the DDPG training phase, for the first \num{50} episodes we add zero-mean Gaussian noise to the actions with \SI{15}{\percent} of the control range as standard deviation.
As for the next \num{50} episodes we reduce the standard deviation to \SI{10}{\percent} of the control range.
Finally, we further reduce the standard deviation to \SI{5}{\percent} of the control range for the remaining episodes.

For every \num{1000} time steps, we update the parameters for \num{1000} $(s, a, r, s')$ tuples in the replay buffer with batch size \num{1}, after which the soft target policy and critic is also updated.
The transitions are selected randomly, but we ensure that at least \SI{5}{\percent} of the transitions are from the newest episode.

For the map creation phase, we use eight DDPG agents in parallel, which update the same map synchronously.
Four of the agents use $I_{\mathrm{init}} = \num{10000}$, while the remaining four agents have $I_{\mathrm{init}}$ set to \num{20000}.
As described in~\Cref{ssec:mmprl}, the agents start selecting random policies from the map to update when they reach $I_{\mathrm{init}}$ iterations.
After the map updates \num{61000} times for the hexapod and \num{321000} times for the Walker2D,
we forcefully stop the map creation phase because the number of the policies in the map converges.
%
For Bayesian optimization, we use parameters $\kappa = \num{0.05}$, the termination condition's hyper parameter $\alpha = \num{0.95}$, the hyper parameter for kernel function $\rho = \num{0.3}$, and $\sigma_{\textrm{noise}} = \num{0.001}$.

\subsection{Baseline Comparison}
We evaluate MMPRL on two aspects: its map-creation ability and adaptability.

\subsubsection{Map-creation}
For evaluating the map-creation ability, we compare MMPRL to the default MAP-Elites algorithm~\cite{cully2014robots}.
More specifically, the speed with which both algorithms generate and store various policies in the map are compared to each other.
Moreover, the quality of the stored policies are compared, where quality is defined in terms of covered distance.
Furthermore, we choose the Walker2D as the agent for comparison because the policy network of the hexapod is too high-dimensional (although the network for Walker2D is still high-dimensional, but less), for which MAP-Elites is not suitable.
The policy generator used in MAP-Elites uses the same MLP as chosen for the actor network in MMPRL (see ~\Cref{ssec:hyperparameters}).

As for the settings of MAP-Elites, because the performance can differ even if the same policy is used, we use the same policy for 10 episodes and update the map in every episode.
During the random sampling stage, each parameter is randomly sampled between [-5, 5].
The decision to use this range specifically was based on the range of network parameters of well learned DDPG policies.
After the first 4000 iterations, a policy is picked from the map after which each variable is perturbed with Gaussian noise $\mathcal{N}(0, 0.1)$.

\subsubsection{Adaptability}
To evaluate the adaptability of our method, we compare MMPRL policies before and after adaptation to single-policy DDPG (without map creation) by measuring the covered distance in different situations.
The DDPG policies are created by running \num{20000} episodes for the hexapod and \num{30000} episodes for Walker2D.
For each model, we run eight agents with different random seeds.
The actor, critic, and hyper parameters are identical to what we use in MMPRL for DDPG agents.
However, we now select the policy which accumulates the most rewards as the single best policy, rather than the one found at the end of training.


\section{Results}
\label{sec:results}
\subsection{Map Creation Phase}
It took about two weeks to create a map with MMPRL for the hexapod and nine days for the Walker2D.

\begin{figure}
	\centering
	\subfigure{\includegraphics[width=0.95\columnwidth]{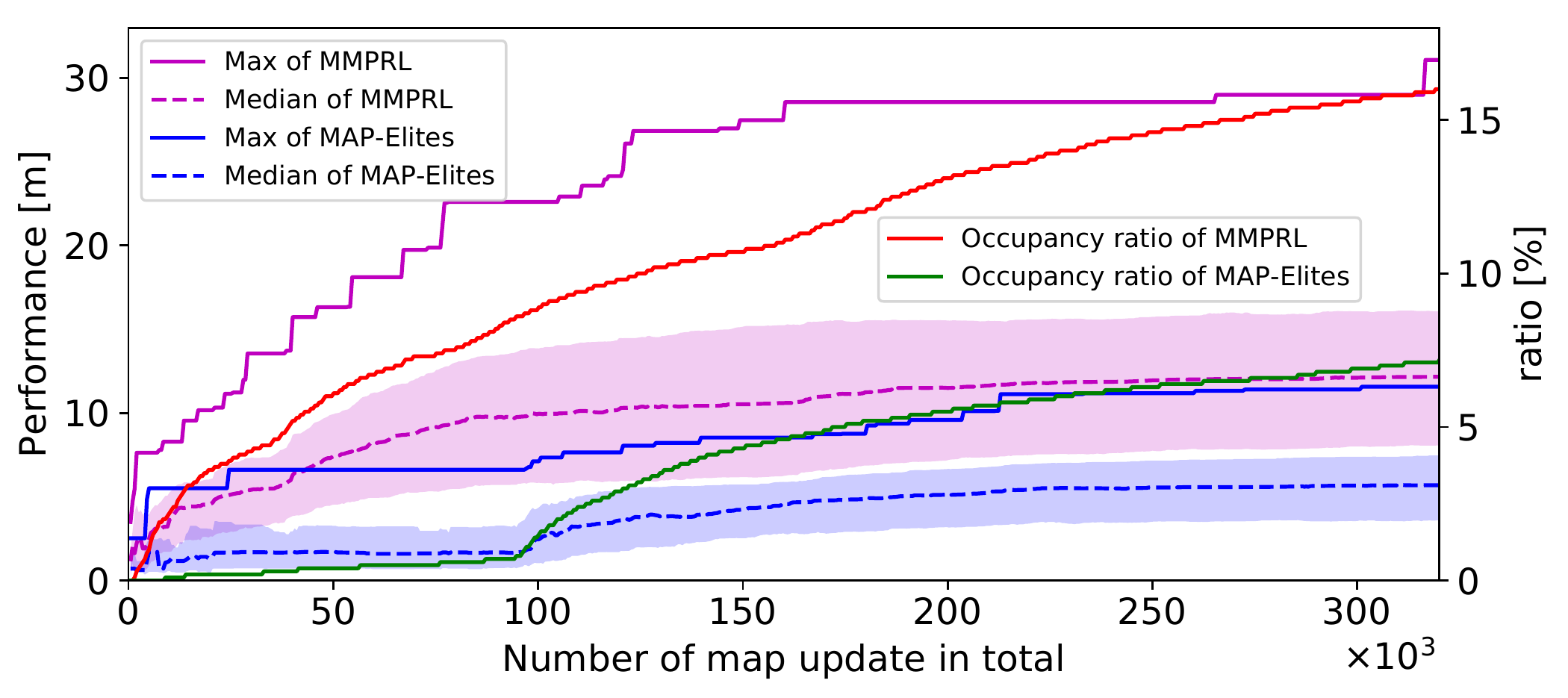}}

\caption{Experiment results of MMPRL and MAP-Elites for the Walker2D during the map creation phase, showing the occupancy ratio and performance as a function of total number of map updates.
	The shaded area shows the 25 and 75 percentiles of the stored policies' performance.}
	\label{fig:map}
	\vspace{-2mm}
\end{figure}

\Cref{fig:map} shows the results of the map creation phase of MMPRL compared with MAP-Elites for the Walker2D experiment.
MMPRL stored \num{2505} policies, which is \SI{16.0}{\percent} of the whole map.
The average performance of the map is \SI{12.4}{\metre} and the standard deviation is \SI{6.0} {\metre}, and the best performing policy covers \SI{31.1}{\metre} in distance.
This shows that our proposed method outperforms MAP-Elites in the number of policies, max and average covered distance.
The policies contain \num{3306} parameters, which is about \num{100} times more than in~\cite{cully2014robots}.
As mentioned in~\cite{cully2014robots} and~\Cref{ss:neuro_evolution}, MAP-Elites and neuroevolutional methods in general do not work well with a high number of parameters such as in our case.

The map generated by MMPRL for the hexapod contains \num{9030} stored policies, which is \SI{57.8}{\percent} of the whole map.
The average performance is \SI{4.0}{\metre} and the standard deviation is \SI{3.0}{\metre}, while the best coverage was \SI{25.5}{\metre}.

Note that the high variances are due to the fact that we not only store the optimal policy, but many policies which display highly different behaviors.
These policies may not be optimal in the default case, but might perform better when a change in situation occurs.

\begin{figure}
\centering
	\vspace{1.5mm}
	\subfigure{\includegraphics[width=0.97\columnwidth]{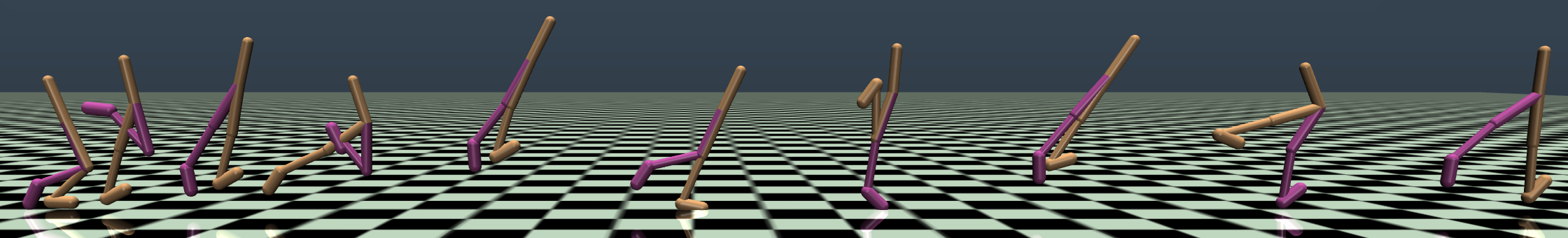}}
    \subfigure{\includegraphics[width=0.97\columnwidth]{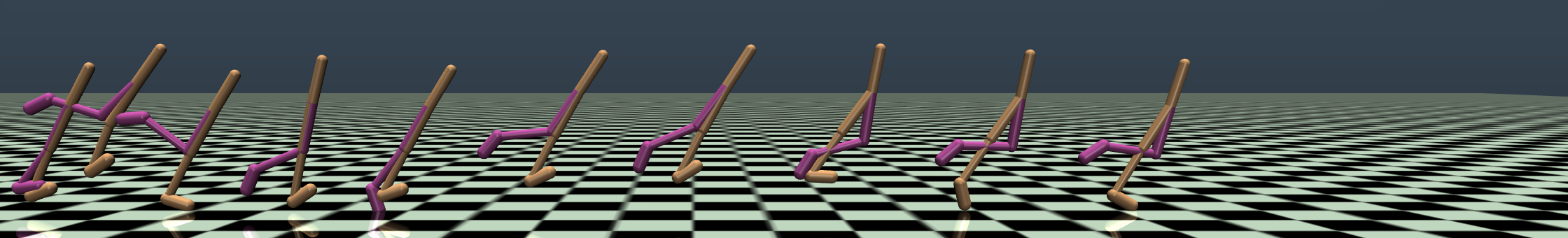}}
	\subfigure{\includegraphics[width=0.97\columnwidth]{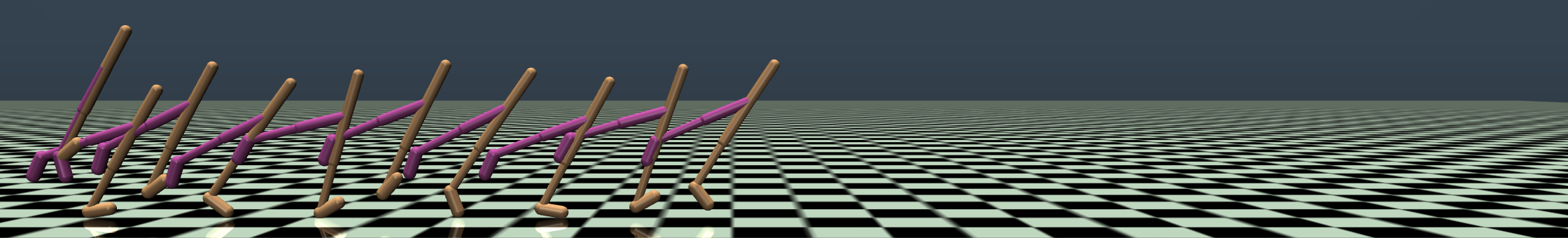}}
	\caption{Time lapse of the Walker2D model walking using different stored policies from time steps 50 to 500.
	The top figure depicts the best policy stored and is used as initial policy for experiments in our adaptation phase, while the other two figures show peculiar behavior by either not making use of joint 4 (middle), or not using the purple leg at all (bottom).}
	\label{fig:walkerexample}
	\vspace{-6mm}
\end{figure}

\Cref{fig:walkerexample} shows some examples of the results by using the stored walking policies.
It shows that the map stores not only very high-performing policies like in the top figure, but also stores the policies with different characteristics, such as in the middle figure where joint 4 is not used, or the bottom figure which rarely uses the left leg (purple).

\begin{figure*}
\centering
\subfigure{\includegraphics[width=1.8\columnwidth]{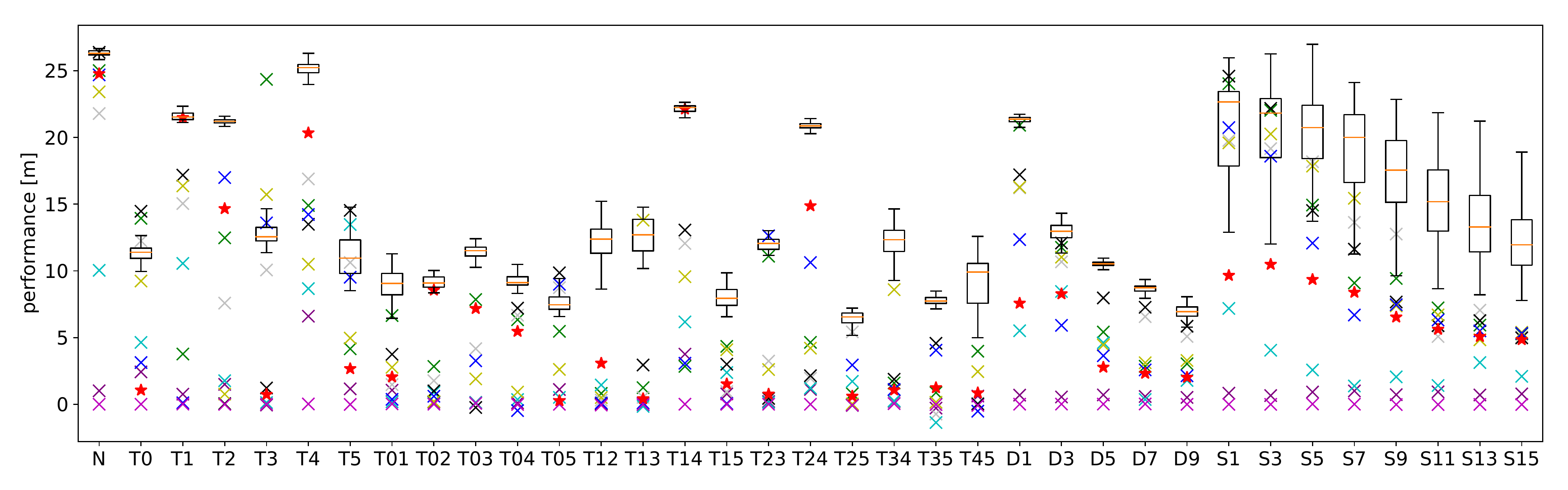}}
\subfigure{\includegraphics[width=1.8\columnwidth]{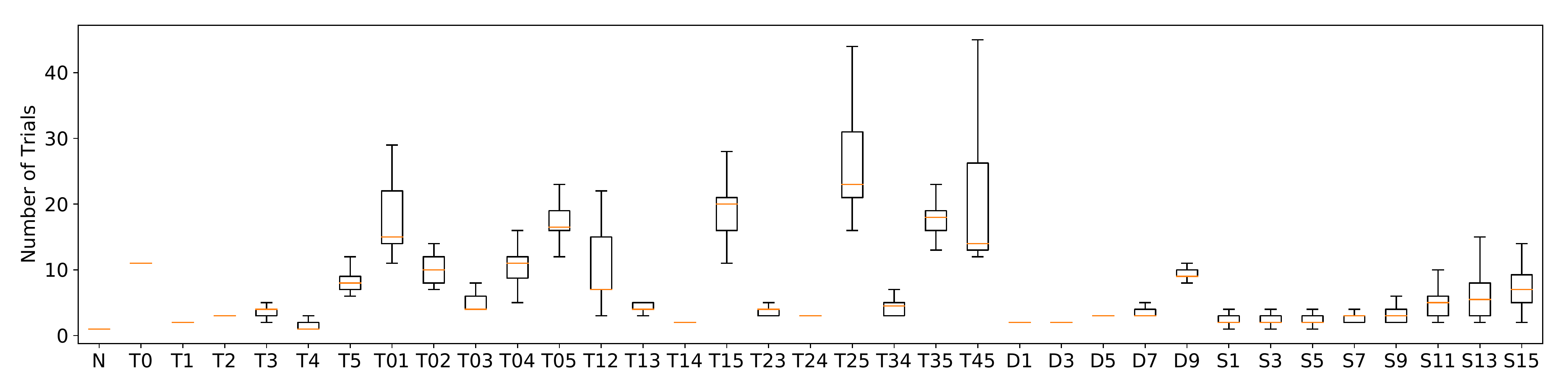}}
	\caption{Results of hexapod experiments showing the performance of the policies of standalone DDPG and post-adaptation MMPRL (top) and the number of trials needed to find a good policy (bottom).
	The horizontal axis show the different scenarios to test the adaptation phase, see~\Cref{ssec:environments} with N being the default situation.
	Scenarios of toe replacements are denoted by T$x$ and T$xy$, where $x$ and $y$ stand the corresponding leg number.
	D$t$ denote experiments where observation delay of $t$ time steps are introduced, while S$a$ stands for the climbing stairs experiments.
	The height of the stairs is a random variable $\mathcal{H}\sim\mathcal{U}(-a, a)$.
	In the top figure, stars represent the initial policies of our map, boxplots show post-adaptation MMPRL policies, and crosses are the eight different standalone DDPG policies (baseline).
	In the bottom figure, the number of trials to find a good policy for a given scenario is depicted by boxplots.
	The boxes show the first and third quartiles, whereas the whiskers show minimum and maximum values within 1.5~IQR.
	}
\label{fig:hexapod}
	\vspace{-5mm}
\end{figure*}

\begin{figure}
\centering
\subfigure{\includegraphics[width=0.9\columnwidth]{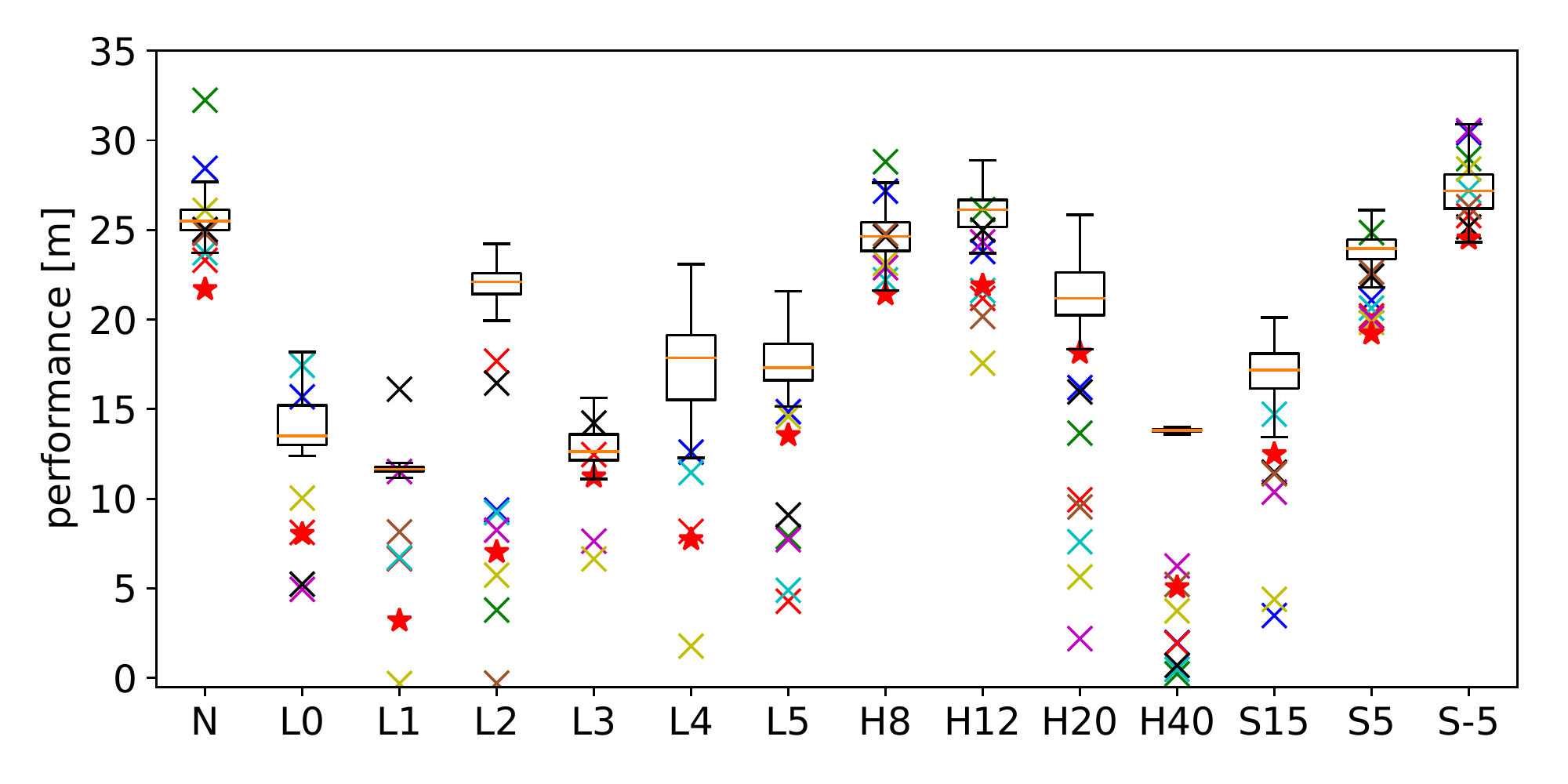}}
\subfigure{\includegraphics[width=0.9\columnwidth]{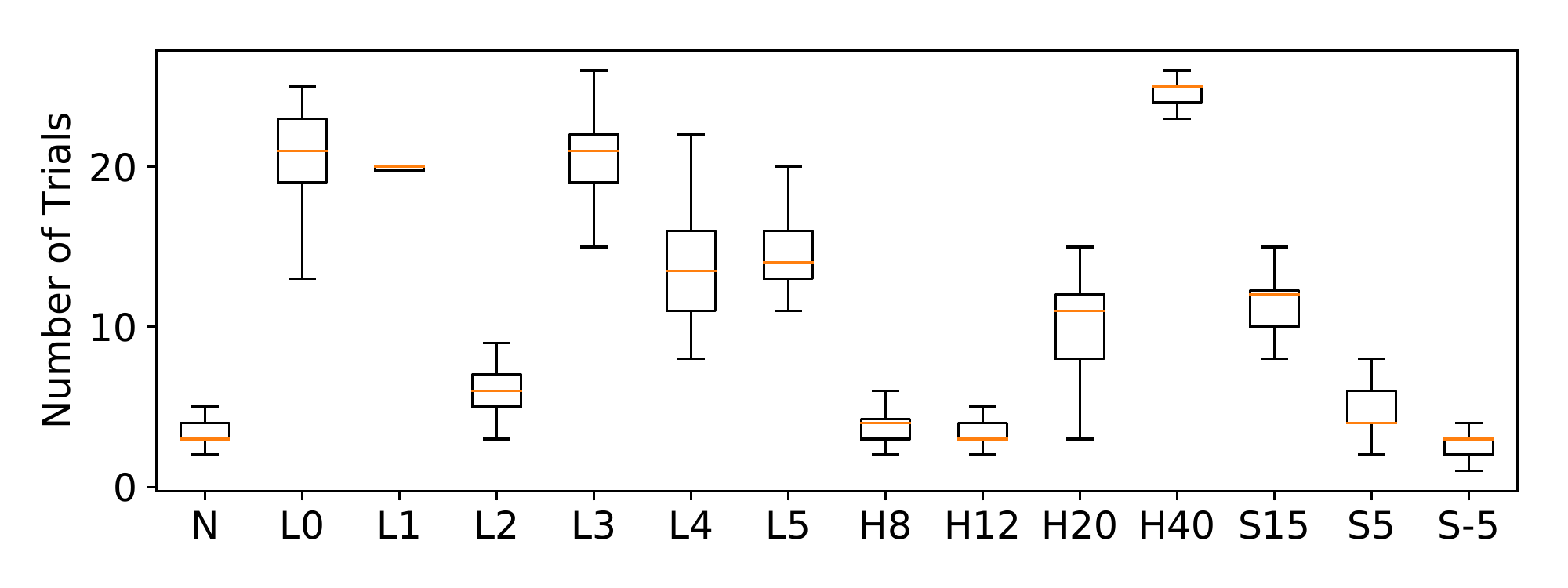}}
	\caption{Results of the Walker2D experiments showing the performance of the policies of standalone DDPG and post-adaptation MMPRL (top) and the number of trials needed to find a good policy (bottom).
	The horizontal axis show the different cases to test the adaptation phase, see~\Cref{ssec:environments} with N being the default situation.
	Cases of a limited joint are denoted by L$j$ where $j$ is the number of the joint being limited.
	H$l$ denote experiments where the head size was changed according to  $\frac{l}{10}\cdot0.05$ where $l \in \mathbb{R}$, while S$d$ stands for the experiments with a simulated slope of $d$ degrees.
	In the top figure, stars represent the initial policies of our map, boxplots show post-adaptation MMPRL policies, and crosses are the eight different standalone DDPG policies (baseline).
	Points with performance below zero have been omitted for some of the cases.
	In the bottom figure, the number of trials to find a good policy for a given scenario is depicted by boxplots.
	The boxes show the first and third quartiles, whereas the whiskers show minimum and maximum values within 1.5~IQR.
	}
\label{fig:walker2d}
	\vspace{-5mm}
\end{figure}

\subsection{Adaptation Phase}
\Cref{fig:hexapod} shows the results of the hexapod experiments and \Cref{fig:walker2d} shows the results of the Walker2D experiments.
Because the initial states are subject to random noise, the performance and the number of trials vary, so we show the average (and quartiles for MMPRL) over \num{100} runs.
The top parts of~\Cref{fig:hexapod} and~\Cref{fig:walker2d} show that MMPRL improves the performance after several trials for both agent models, even with a low-performing initial policy.
The middle subfigure of~\Cref{fig:walkerexample} is an example where the agent found a policy without using joint 4.
During the adaptation phase we observe that this policy is indeed selected often when we disable joint 4 (T4 in~\Cref{fig:hexapod}).
For many cases, the performance of our policies after adaptation is better than single DDPG policies.
This is especially true in difficult situations when the robot loses toes from adjacent legs (e.g., T03, T12, T34 in~\Cref{fig:hexapod}) or climbs steps which are higher than during training (e.g., S15 in~\Cref{fig:hexapod}).

Furthermore, some changes that we introduce to test adaptability (see~\Cref{ssec:environments}) are not directly related to the behavioral descriptor.
Examples of these are the introduction of a delayed observation (D in~\Cref{fig:hexapod}) or when simulating a sloped environment (S in~\Cref{fig:walker2d}).
However, our method can still find a good policy, even in these cases.

\begin{figure}
	\centering
	\subfigure{\includegraphics[width=0.95\columnwidth]{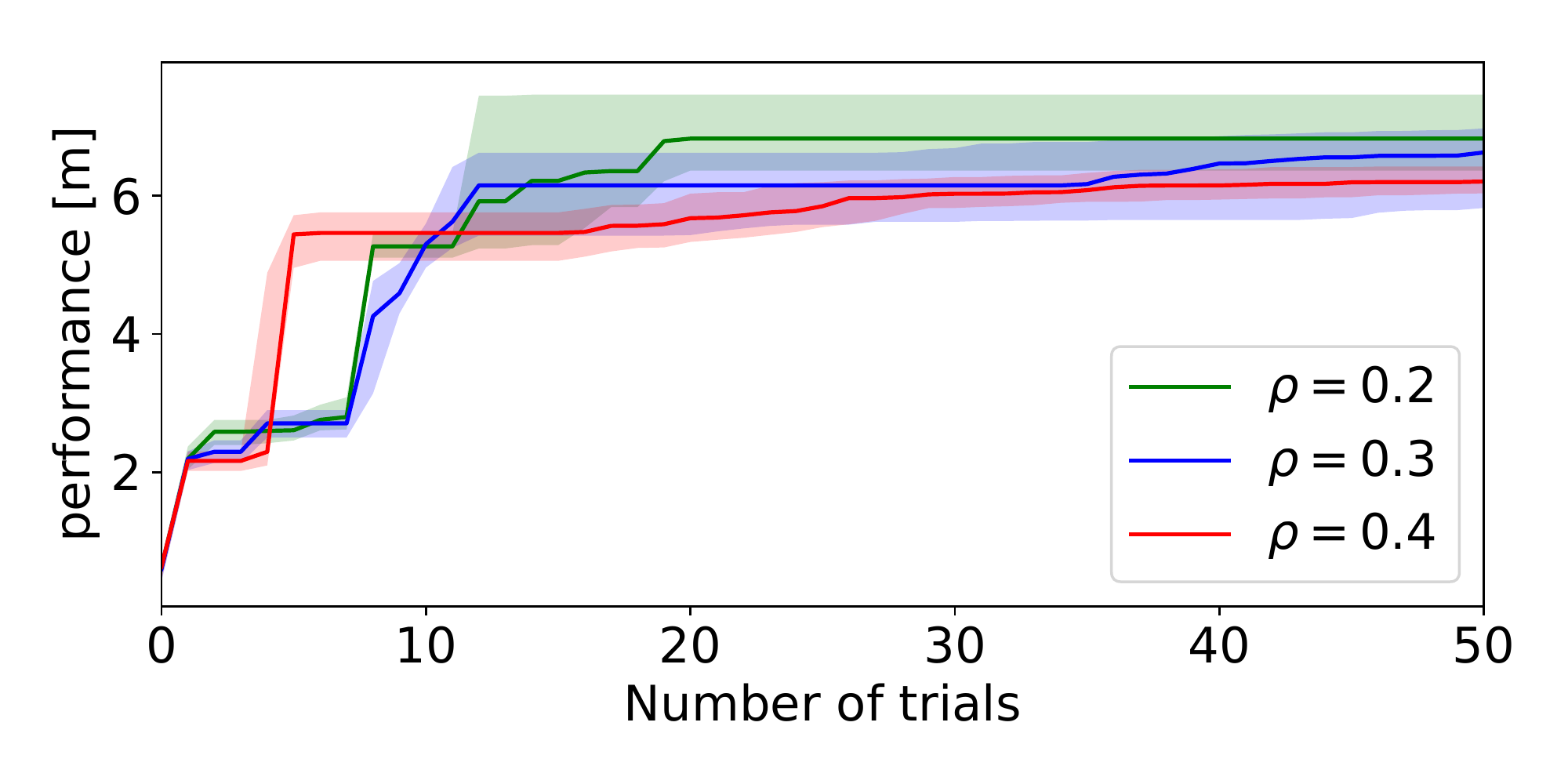}}
    \subfigure{\includegraphics[width=0.95\columnwidth]{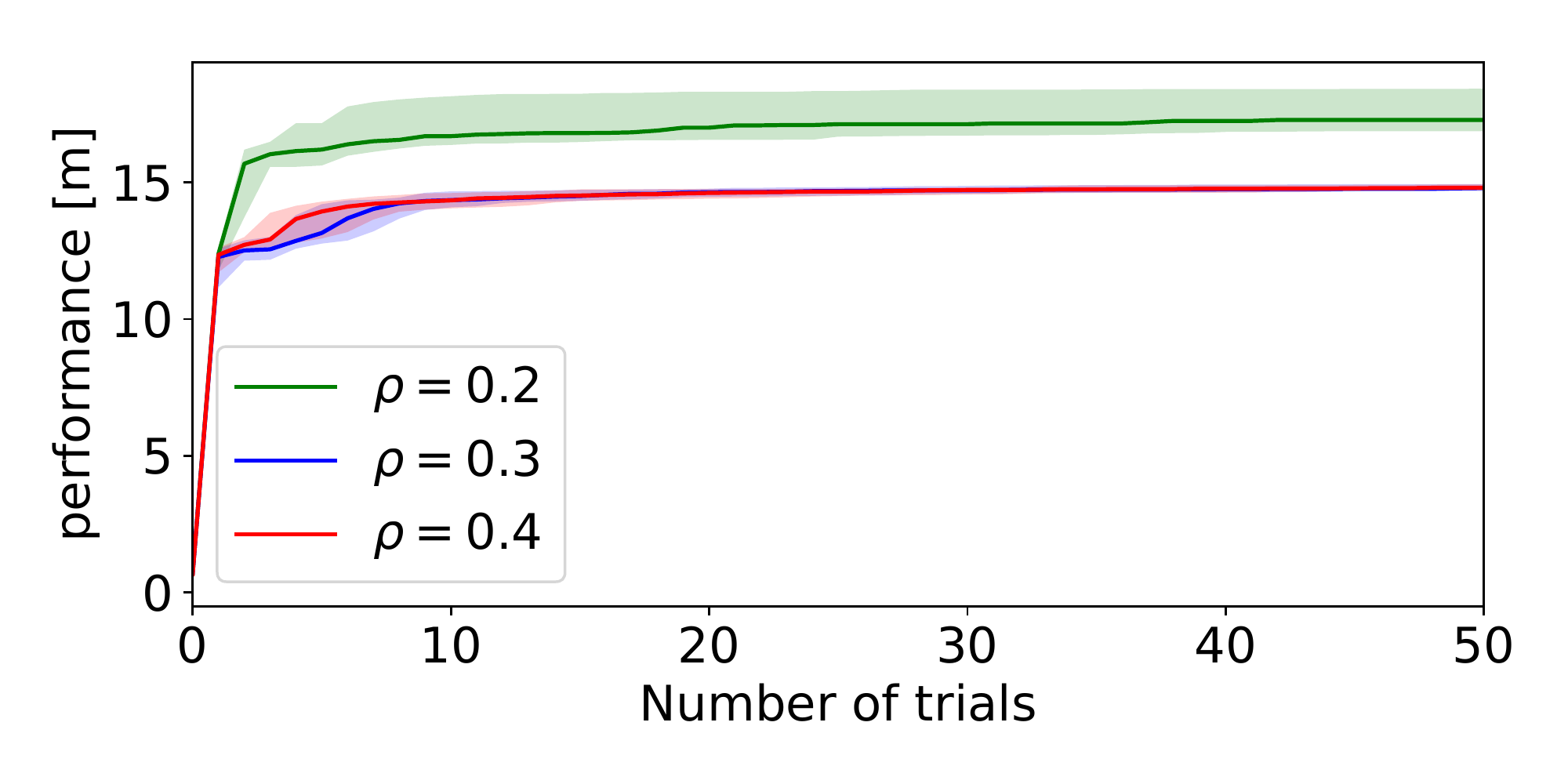}}
	\vspace{-2mm}
	\caption{Results showing the performance during adaptation for cases in which the hexapod is missing toes for legs 2 and 5 (top), and only leg 3 (bottom). The adaptation phase is done without a stopping criterion for up to 50 trials using three different $\rho$ values. Solid lines show mean performances in 100 runs, while the 25 and 75 percentiles are the shaded area.}
	\label{fig:rho}
	\vspace{-5mm}
\end{figure}

The lower parts of~\Cref{fig:hexapod} and~\Cref{fig:walker2d} show that our method adapts within a low amount of trials, which means that adaptation is done within realistic time.
For example, whenever the robot loses one toe, the average number of trials is less than \num{10}.
For T3 in~\Cref{fig:hexapod}, the average number of trials taken over \num{100} runs is \num{3.5} and it takes \SI{11.4}{\second} on our machine, which are \SI{175}{\second} in simulation time.
In T25, which is one of the slowest cases in our experiments, the mean number of trials in \num{100} runs is \num{27.3} and takes \SI{90.1}{\second}, \SI{1365}{\second} in simulation time.
We believe this time is acceptable for use in actual environments.
The duration of the adaptation phase can be controlled by changing the parameters of M-BOA.
By decreasing $\rho$ or increasing $\alpha$, a better policy may be found, but may need more trials like for T25 in~\Cref{fig:hexapod}.
\Cref{fig:rho} shows the progress of the adaptation phase for different $\rho$ values.
We see that $\rho = 0.2$ performs best in both scenarios T25 and T3.
However, it took many trials to converge in the more difficult case T25, while $\rho = 0.4$ found better policies than $\rho = 0.2$ or $\rho = 0.3$ when adaptation must be done in between at most 5 to 10 trials.

In several rare cases, the single DDPG policies perform better than our policies after adaptation (e.g., T3 in~\Cref{fig:hexapod} or H8 and S-5 in~\Cref{fig:walker2d}).
Two possible reasons are that
\begin{inparaenum}[(i)]
	\item M-BOA could not find the best policy in the new situation, or
	\item the single best policy of one DDPG agent is better than the best policy in our map.
\end{inparaenum}
In the case of (i), a better policy may be found by changing $\rho$ or $\alpha$ of M-BOA.
For example if we change $\rho$ to \num{0.2}, MMPRL will have similar performance as the second best DDPG policy, as depicted in \Cref{fig:rho}.
Case (ii) is a possibility because the policies of MMPRL are also generated by DDPG.
The difference is that some policies in the map may not have converged to the optimal policy, because they have not been updated enough to converge.
Therefore it is only natural that the optimal policy found by DDPG works better in at least the normal situation.
However, we see in~\Cref{fig:hexapod} and~\Cref{fig:walker2d} that the same standalone DDPG policies never perform well in different situations.
Hence, we conclude that a single policy is not robust enough to adapt to situation changes, while MMPRL can find policies for various conditions.

\section{Conclusion}
\label{sec:conclusion}
We proposed and implemented MMPRL to enhance the adaptability of robots.
The results from our experiments show enhanced adaptability for both the Walker2D and the hexapod agents to various environmental changes and body injuries compared to single-policy DDPG.
Although some standalone DDPG policies occasionally perform better than one of the chosen policies by MMPRL after adaptation, they never perform well in different scenarios.
Moreover, we showed that our method is able to aqcuire higher-dimensional policies for the Walker2D in a more efficient and effective manner compared to the default MAP-Elites algorithm.

For future work, we note that although the current map creation method requires a lot of time, it could be optimized by encouraging the search of undiscovered states by adding a curiosity term to the reward.
Finally, we plan to extend our method and show its effectiveness by applying it to physical robots through e.g., transfer learning.
In particular, we will extend this work for humanoid robots, which are known to be less tolerant to complex environments.

\section*{Acknowledgements}
The authors would like to thank Shin-ichi Maeda for his assistance and support on this work.

\bibliographystyle{IEEEtran}
\bibliography{IEEEabrv,bibliography}

\begin{thebibliography}{10}
\providecommand{\url}[1]{#1}
\csname url@rmstyle\endcsname
\providecommand{\newblock}{\relax}
\providecommand{\bibinfo}[2]{#2}
\providecommand\BIBentrySTDinterwordspacing{\spaceskip=0pt\relax}
\providecommand\BIBentryALTinterwordstretchfactor{4}
\providecommand\BIBentryALTinterwordspacing{\spaceskip=\fontdimen2\font plus
\BIBentryALTinterwordstretchfactor\fontdimen3\font minus
  \fontdimen4\font\relax}
\providecommand\BIBforeignlanguage[2]{{%
\expandafter\ifx\csname l@#1\endcsname\relax
\typeout{** WARNING: IEEEtran.bst: No hyphenation pattern has been}%
\typeout{** loaded for the language `#1'. Using the pattern for}%
\typeout{** the default language instead.}%
\else
\language=\csname l@#1\endcsname
\fi
#2}}

\bibitem{lillicrap2015continuous}
T.~P. Lillicrap, \emph{et~al.}, ``Continuous control with deep reinforcement
  learning,'' \emph{arXiv preprint arXiv:1509.02971}, 2015.

\bibitem{mnih2015human}
V.~Mnih, \emph{et~al.}, ``Human-level control through deep reinforcement
  learning,'' \emph{Nature}, vol. 518, no. 7540, pp. 529--533, 2015.

\bibitem{pinto2017robust}
L.~Pinto, \emph{et~al.}, ``Robust adversarial reinforcement learning,'' 2017.

\bibitem{cully2014robots}
A.~Cully, \emph{et~al.}, ``{Robots that can adapt like animals},''
  \emph{Nature}, vol. 521, pp. 503--507, 2015.

\bibitem{chatzilygeroudisreset}
K.~Chatzilygeroudis, \emph{et~al.}, ``Reset-free trial-and-error learning for
  data-efficient robot damage recovery,'' \emph{arXiv preprint
  arXiv:1610.04213}, 2016.

\bibitem{bongard2006resilient}
J.~Bongard, \emph{et~al.}, ``Resilient machines through continuous
  self-modeling,'' \emph{Science}, vol. 314, no. 5802, pp. 1118--1121, 2006.

\bibitem{koos2013fast}
S.~Koos, \emph{et~al.}, ``Fast damage recovery in robotics with the
  t-resilience algorithm,'' \emph{The International Journal of Robotics
  Research}, vol.~32, no.~14, pp. 1700--1723, 2013.

\bibitem{ren2015multiple}
G.~Ren, \emph{et~al.}, ``Multiple chaotic central pattern generators with
  learning for legged locomotion and malfunction compensation,''
  \emph{Information Sciences}, vol. 294, pp. 666--682, 2015.

\bibitem{mnih2016asynchronous}
V.~Mnih, \emph{et~al.}, ``Asynchronous methods for deep reinforcement
  learning,'' in \emph{Proc. ICML}, 2016.

\bibitem{schulman2015trust}
J.~Schulman, \emph{et~al.}, ``Trust region policy optimization,'' in
  \emph{Proc. ICML}, 2015.

\bibitem{levine2016end}
S.~Levine, \emph{et~al.}, ``End-to-end training of deep visuomotor policies,''
  \emph{Journal of Machine Learning Research}, vol.~17, no.~39, pp. 1--40,
  2016.

\bibitem{li2016nlp}
J.~Li, \emph{et~al.}, ``Deep reinforcement learning for dialogue generation,''
  in \emph{Proc. EMNLP}, 2016.

\bibitem{zoph2017}
L.~Zoph, ``Neural architecture search with reinforcement learning,''
  \emph{Proc. ICLR}, 2017.

\bibitem{li2017deep}
Y.~Li, ``Deep reinforcement learning: An overview,'' \emph{arXiv preprint
  arXiv:1701.07274}, 2017.

\bibitem{yu2017preparing}
W.~Yu, \emph{et~al.}, ``Preparing for the unknown: Learning a universal policy
  with online system identification,'' \emph{arXiv preprint arXiv:1702.02453},
  2017.

\bibitem{Heess2015}
N.~Heess, \emph{et~al.}, ``{Memory-based control with recurrent neural
  networks},'' in \emph{Workshop on Deep Reinforcement Learning in NIPS}, 2015.

\bibitem{higuera2017adapting}
J.~C.~G. Higuera, \emph{et~al.}, ``Adapting learned robotics behaviours through
  policy adjustment,'' in \emph{Proc. IEEE ICRA}, 2017.

\bibitem{williams2017information}
G.~Williams, \emph{et~al.}, ``Information theoretic mpc for model-based
  reinforcement learning,'' in \emph{Proc. IEEE ICRA}, 2017.

\bibitem{tobin2017domain}
J.~Tobin, \emph{et~al.}, ``Domain randomization for transferring deep neural
  networks from simulation to the real world,'' \emph{arXiv preprint
  arXiv:1703.06907}, 2017.

\bibitem{rajeswaran2016epopt}
A.~Rajeswaran, \emph{et~al.}, ``Epopt: Learning robust neural network policies
  using model ensembles,'' \emph{Proc. ICLR}, 2017.

\bibitem{pinto2017supervision}
L.~Pinto, \emph{et~al.}, ``Supervision via competition: Robot adversaries for
  learning tasks,'' in \emph{Proc. IEEE ICRA}, 2017.

\bibitem{pmlr-v70-vezhnevets17a}
A.~S. Vezhnevets, \emph{et~al.}, ``Feudal networks for hierarchical
  reinforcement learning,'' in \emph{Proc. ICML}, 2017.

\bibitem{yao1993review}
X.~Yao, ``A review of evolutionary artificial neural networks,''
  \emph{International journal of intelligent systems}, vol.~8, no.~4, pp.
  539--567, 1993.

\bibitem{angeline1994evolutionary}
P.~J. Angeline, \emph{et~al.}, ``An evolutionary algorithm that constructs
  recurrent neural networks,'' \emph{IEEE Transactions on Neural Networks},
  vol.~5, no.~1, pp. 54--65, 1994.

\bibitem{yao1999evolving}
X.~Yao, ``Evolving artificial neural networks,'' \emph{Proceedings of the
  IEEE}, vol.~87, no.~9, pp. 1423--1447, 1999.

\bibitem{miikkulainen2017evolving}
R.~Miikkulainen, \emph{et~al.}, ``Evolving deep neural networks,'' \emph{arXiv
  preprint arXiv:1703.00548}, 2017.

\bibitem{alvernaz2017autoencoder}
S.~Alvernaz and J.~Togelius, ``Autoencoder-augmented neuroevolution for visual
  doom playing,'' \emph{arXiv preprint arXiv:1707.03902}, 2017.

\bibitem{salimans2017evolution}
T.~Salimans, \emph{et~al.}, ``Evolution strategies as a scalable alternative to
  reinforcement learning,'' \emph{arXiv preprint arXiv:1703.03864}, 2017.

\bibitem{watkins1992q}
C.~J. Watkins and P.~Dayan, ``Q-learning,'' \emph{Machine learning}, vol.~8,
  no. 3-4, pp. 279--292, 1992.

\bibitem{mnih2013playing}
V.~Mnih, \emph{et~al.}, ``Playing atari with deep reinforcement learning,''
  \emph{arXiv preprint arXiv:1312.5602}, 2013.

\bibitem{mockus2012bayesian}
J.~Mockus, \emph{Bayesian approach to global optimization: theory and
  applications}.\hskip 1em plus 0.5em minus 0.4em\relax Springer Science \&
  Business Media, 2012, vol.~37.

\bibitem{islam2017reproducibility}
R.~Islam, \emph{et~al.}, ``{Reproducibility of benchmarked deep reinforcement
  learning tasks for continuous control},'' in \emph{Workshop on Deep
  Reinforcement Learning in ICML}, 2017.

\bibitem{1606.01540}
G.~Brockman, \emph{et~al.}, ``Openai gym,'' \emph{arXiv preprint
  arXiv:1606.01540}, 2016.

\bibitem{todorov2012mujoco}
E.~Todorov, \emph{et~al.}, ``Mujoco: A physics engine for model-based
  control,'' in \emph{Proc. IEEE IROS}, 2012.

\bibitem{chainer_learningsys2015}
S.~Tokui, \emph{et~al.}, ``{Chainer: a next-generation open source framework
  for deep learning},'' in \emph{Workshop on Machine Learning Systems in NIPS},
  2015.

\bibitem{hexapod_model}
\BIBentryALTinterwordspacing
"phantomx\_description". [Online]. Available:
  \url{https://github.com/HumaRobotics/phantomx_description}
\BIBentrySTDinterwordspacing

\bibitem{kingma2014adam}
D.~Kingma and J.~Ba., ``Adam: A method for stochastic optimization,''
  \emph{Proc. ICLR}, 2015.

\end{thebibliography}

\end{document}